\documentclass[a4paper,nocompress,11pt]{article}
\usepackage{subcaption}
\usepackage{graphicx}
\usepackage{xcolor}
\usepackage{amsmath}
\usepackage{amsfonts}
\usepackage{amssymb}
\usepackage{booktabs}
\usepackage{threeparttable}
\usepackage{array}
\usepackage{tabularx}
\usepackage{multirow}
\usepackage{float}
\usepackage{geometry}
\definecolor{panelbg}{HTML}{FBF7EF}
\definecolor{panelframe}{HTML}{D8D2C8}

\title{Magnification-Invariant Image Classification via Domain Generalization and Stable Sparse Embedding Signatures}
\author{Ifeanyi Ezuma, Olusiji Medaiyese}
\date{}
\begin{document}
\maketitle

\begin{abstract}
\noindent Magnification shift is a major obstacle to robust histopathology classification, because models trained on one imaging scale often generalize poorly to another. Here, we evaluated this problem on the BreaKHis dataset using a strict patient-disjoint leave-one-magnification-out protocol, comparing supervised baseline, baseline augmented with DCGAN-generated patches, and a gradient-reversal domain-general model designed to preserve discriminative information while suppressing magnification-specific variation. Across held-out magnifications, the domain-general model achieved the strongest overall discrimination and its clearest gain was observed when 200X was held out. By contrast, GAN augmentation produced inconsistent effects, improving some folds but degrading others, particularly at 400X. The domain-general model also yielded the lowest Brier score at 0.063 vs 0.089 at baseline. Sparse embedding analysis further revealed that domain-general training reduced average signature size more than three-fold (306 versus 1,074 dimensions) while preserving equivalent predictive performance (AUC: 0.967 vs 0.965; F1: 0.930 vs 0.931). It also increased cross-fold signature reproducibility from near-zero Jaccard overlap in the baseline to 0.99 between the 100X and 200X folds. These findings show that calibrated, compact, and transferable representations can be learned without added architectural complexity, with clear implications for the reliable deployment of computational pathology models across heterogeneous acquisition settings.
\end{abstract}

\section*{I. Introduction}
\noindent Histopathology remains central to breast cancer diagnosis, and one of the key tasks is distinguishing benign from malignant tumours. Benign tumours are non-cancerous growths, whereas malignant tumours are cancerous and are often further described as non-invasive (in situ) or invasive, depending on whether they have penetrated surrounding tissue. The digitization of microscopy slides has accelerated the push toward computational tools that can support pathologists at scale. Deep learning has become a leading approach for these images because it can learn multi-scale morphological patterns directly from data [1,2], often reducing reliance on hand-engineered descriptors. However, translating strong in-distribution performance into reliable clinical behavior is difficult: histology pipelines vary across labs and scanners, and even modest differences in acquisition or preparation can induce domain shift that degrades model accuracy when evaluated on unseen conditions. These concerns are now widely recognized in the digital pathology literature as a key barrier to dependable deployment [3,4]. \\

\noindent A particularly concrete, widely used testbed for studying such shift is the BreaKHis dataset, which contains 7,909 breast tumor images from 82 patients captured at four magnifications (40×, 100×, 200×, 400×) with benign/malignant labels and subtype annotations [5]. In practice, magnification changes alter both the visual field-of-view and the apparent texture and structure statistics, meaning that a classifier trained at one magnification can fail to transfer to another. This makes magnification invariance naturally align with the domain generalization setting: learning from one or more source domains to perform well on an unseen target domain. This is an area with a broad methodological landscape in computer vision [6]. 

\noindent Magnification shift has been recognized as a significant challenge in generalization on BreaKHis, leading to methods that learn embeddings transferable across 40×, 100×, 200×, and 400×. In practice, previous studies report both within-magnification testing and cross-magnification testing. The performance gap between these settings has led to specific strategies for magnification-aware or magnification-invariant learning. For instance, Bayramoglu et al. found that magnification-independent CNN training achieved 83.25\% average patient-level recognition. Meanwhile, a multi-task CNN that predicted malignancy and magnification together reached 82.13\%. This suggests that magnification information can be included during training without a major loss in diagnostic performance [7]. Gupta et al. reported 87.53\% average patient-level recognition for a magnification-independent ensemble and about 84.21\% mean cross-magnification accuracy across holdouts for magnification-specific models [8]. In a stricter unseen-domain setting, Sikaroudi et al. treated magnifications as separate domains. Under leave-one-magnification-out evaluation, they reported mean accuracies of 80.06\% for binary malignancy classification and 49.07\% for 8-class tumor-type classification across the four held-out magnifications [9]. Chhipa et al. also used magnification as a training signal in a self-supervised framework. They reported a mean type-2 cross-magnification image-level accuracy of 83.88\% and mean patient-level accuracy of 83.25\% across held-out magnifications for their best variant [10]. \\

\noindent In parallel, generative adversarial networks (GAN) have become a practical tool in digital pathology to handle appearance variability, especially through synthetic augmentation that expands the range of plausible training images [11,12]. However, while such an augmentation can reduce nuisance variation, it does not by itself guaranty that the learned representation is stable and consistently predictive across magnifications. \\

\noindent In this study, we address magnification shift as a central challenge in robust classification on BreaKHis. We combined a strict patient-disjoint leave-one-magnification-out (LOMO) evaluation framework with domain-adversarial representation learning, while also using GAN augmentation to test whether increased training diversity alone improves transfer to unseen magnifications. Specifically, we learn a domain-general representation that preserves discriminative information while reducing magnification-specific variation. We further complement the predictive analysis with a sparse embedding study to assess whether the learned latent space yields a compact and stable signature across LOMO folds. Particularly, we introduce: (i) a LOMO evaluation protocol with patient-level leakage controls, (ii) a Deep Convolutional Generative Adversarial Network (DCGAN) for generating synthetic images to expand domain variability during training, (iii) a magnification-invariant representation learning objective, and (iv) a stability-driven sparse signature extraction procedure.

\section*{2. Dataset}
The BreaKHis dataset, created by Spanhol et al. [5], contains 7,909 histological breast cancer images from 82 patients in a 2014 clinical study. Captured using an Olympus BX-50 microscope and a Samsung Digital Color Camera, the images are categorized as benign or malignant, each with four subtypes. The data set features RGB images with 700×460 pixel resolution, taken at 40×, 100×, 200×, and 400× magnifications. Pathologists identified regions of interest for diagnosis, with extraneous areas removed for consistency. The distribution of the data over the different magnifications for the benign and malignant groups is shown in Table 1. 

\begin{table}[h]
\centering
\caption{BreaKHis image counts by magnification and label.}
\label{tab:breakhis_counts}
\begin{tabular}{lrrr}
\hline
\textbf{Label} & \textbf{Benign} & \textbf{Malignant} & \textbf{Total} \\
\hline
40X  & 625 & 1{,}370 & 1{,}995 \\
100X & 644 & 1{,}437 & 2{,}081 \\
200X & 623 & 1{,}390 & 2{,}013 \\
400X & 588 & 1{,}232 & 1{,}820 \\
\hline
\textbf{Total of Images} & \textbf{2{,}480} & \textbf{5{,}429} & \textbf{7{,}909} \\
\hline
\end{tabular}
\end{table}

\subsection*{2.1. Data Partitioning}
Given that histopathology datasets contain multiple patches per patient, and patches from the same patient can exhibit correlated staining and morphology. This could influence training and evaluation sets, inflating performance estimates due to patient-level leakage. To prevent this, we performed data partitioning at the patient level using stratified group splitting, where the grouping variable is subject ID and stratification is performed on the joint label–magnification strata (i.e., label × magnification) to preserve class balance within each magnification. This yields a patient-level split of roughly 64\%/16\%/20\% for train/validation/test, respectively, with no patient overlap between the three sets. This split serves as the canonical partition for all subsequent experiments in this study. 

\subsection*{2.2. Leave-One-Magnification-Out Evaluation}
To quantify robustness under magnification shift, we adopt a LOMO evaluation protocol, in which each magnification level is treated as an unseen domain at test time. For each held-out magnification $m \in \{40x,100x,200x,400x\}$, the training and validation sets include only samples with magnification $\mathcal{M} \neq m$, while the test set includes only samples with $\mathcal{M}=m$. We enforce strict patient-level leakage control by first constructing a patient-disjoint train/validation/test split and then applying magnification filtering within each split as shown in Figure 2. This yields four out-of-domain evaluations, one per held-out magnification, with consistent patient partitions across folds and fold-specific image counts determined by the availability of patches at each magnification. We evaluated all the methods proposed in this study under the LOMO protocol.

\begin{figure}[h]
    \centering
    \includegraphics[width=0.85\textwidth]{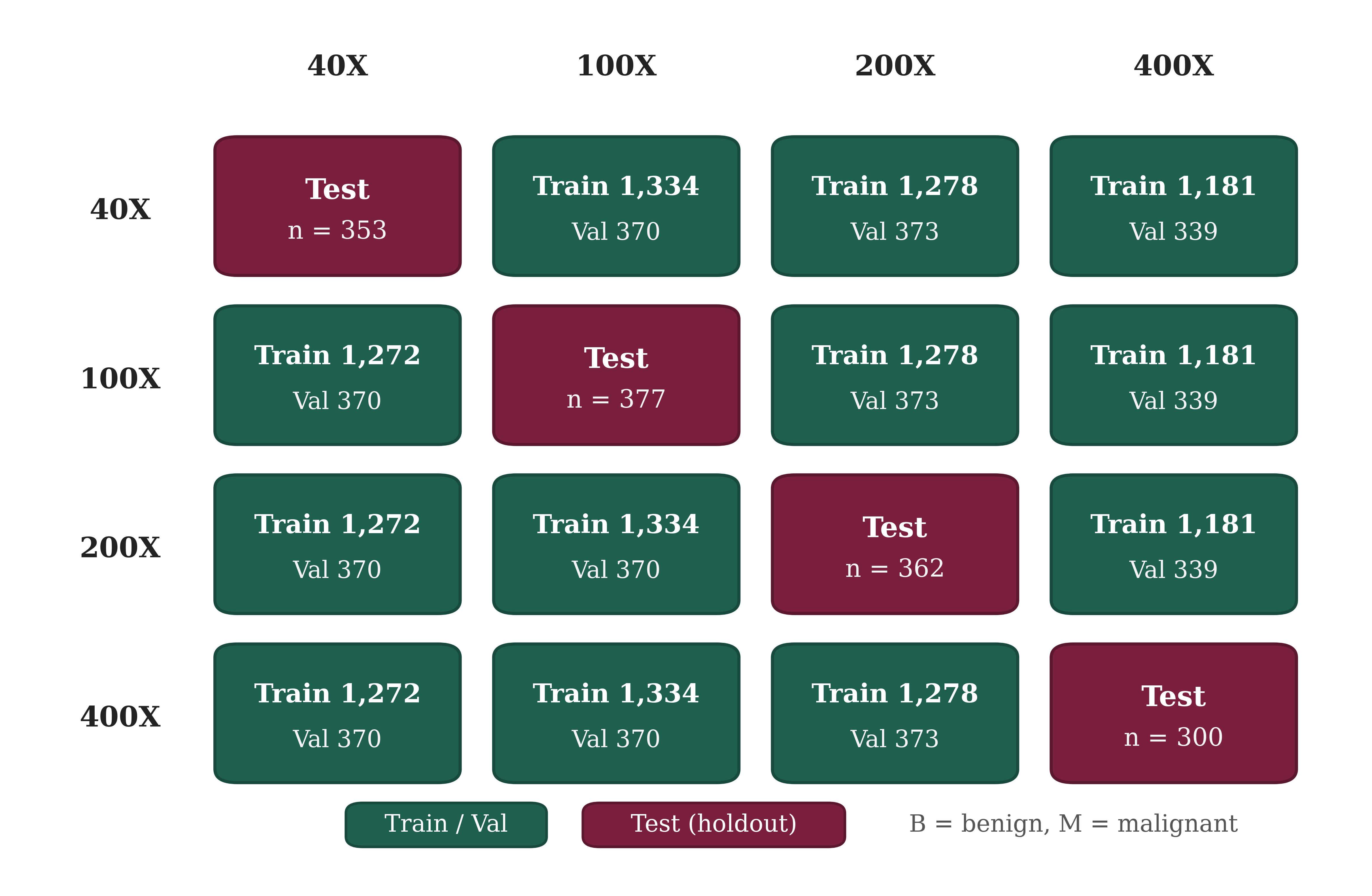}
    \caption{LOMO train/validation/test partitioning across magnifications.}
    \label{fig:model}
\end{figure}

\subsection*{2.3. Preprocessing}
We adopt a pretrained network (ResNet-50) and replace the final classification layer with a binary output head for benign vs malignant prediction. All patches are loaded as RGB images and resized to a fixed input resolution compatible with ResNet-50. For validation and test sets, we apply only deterministic preprocessing (resize/center-crop and normalization) to ensure evaluation stability. We normalize intensities using ImageNet statistics to match the pretrained backbone initialization. Models are trained using mini-batch stochastic optimization with a binary cross-entropy objective. To account for class imbalance, we use a positive-class weighting term computed from the training split of each LOMO fold, and this weight is recomputed separately for each LOMO run to reflect the fold-specific class distribution. Hyperparameters (learning rate, weight decay) are then tuned using the validation split. The final model for each fold is selected based on validation performance and then evaluated on the held-out magnification test set.

\subsection*{2.4. GAN Augmentation}
To increase training diversity, we incorporate data augmentation using DCGAN [13]. We enrich the empirical training distribution with synthetic patches that improve sample diversity and partially mitigate class imbalance. Let $x \sim p_{\text{data}}$ represent a real patch and $z \sim p_z$ a latent noise vector. The generator $G(z)$ maps latent samples to synthetic image patches, while the discriminator $D(x)$ is trianed to distinguish real from generated samples. Our GAN model is optimized with the adversarial objective function [14]: \\
\[
\min_G \max_D \; E_{x\sim p_{\text{data}}} \left[ \log D(x)\right] + 
E_{x\sim p_z} \left[ \log \left(1- D(G(z)) \right) \right]
\]
For each LOMO fold, we train separate class-specific DCGANs using only the real images from the corresponding training partition. This yields one generator for benign patches and one for malignant patches, enabling targeted augmentation within each class while preserving strict separation from validation and test data. Synthetic images are generated at a fixed resolution consistent with ResNet-50, and the number of generated samples is controlled such that real images remain dominant in the final training set. In this study, GAN is treated as an ablation for our pipeline, allowing us to isolate its contribution and compare performance against our proposed method. \\

\begin{figure}[H]
    \centering
    
    \begin{minipage}{0.11\textwidth}
        \centering
        \textbf{Real data}
    \end{minipage}
    \begin{minipage}{0.20\textwidth}
        \centering
        \includegraphics[width=\linewidth]{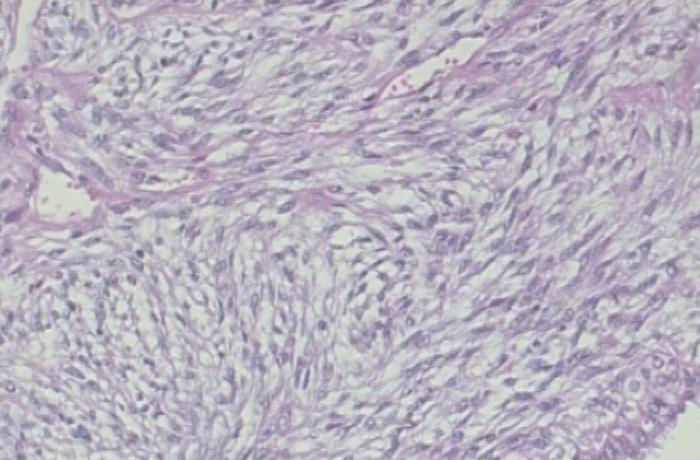}
    \end{minipage}
    \begin{minipage}{0.20\textwidth}
        \centering
        \includegraphics[width=\linewidth]{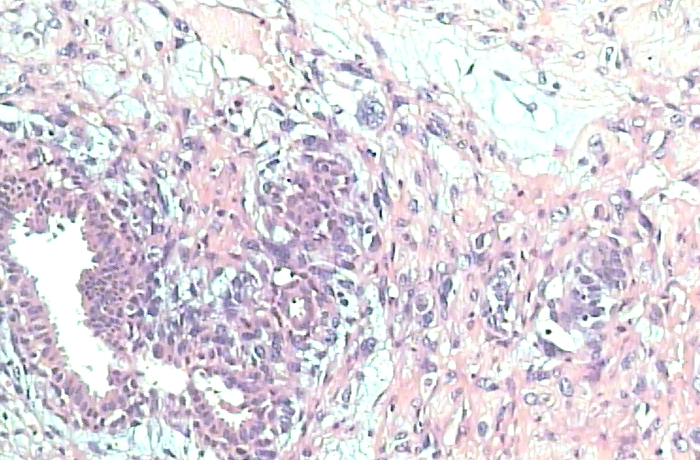}
    \end{minipage}
    \begin{minipage}{0.20\textwidth}
        \centering
        \includegraphics[width=\linewidth]{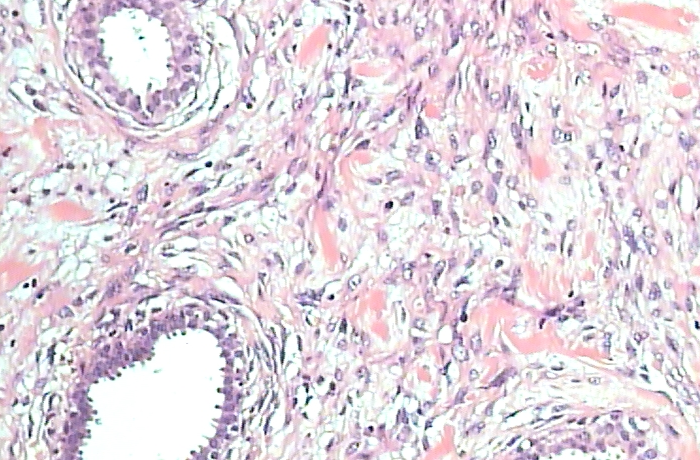}
    \end{minipage}
    \begin{minipage}{0.20\textwidth}
        \centering
        \includegraphics[width=\linewidth]{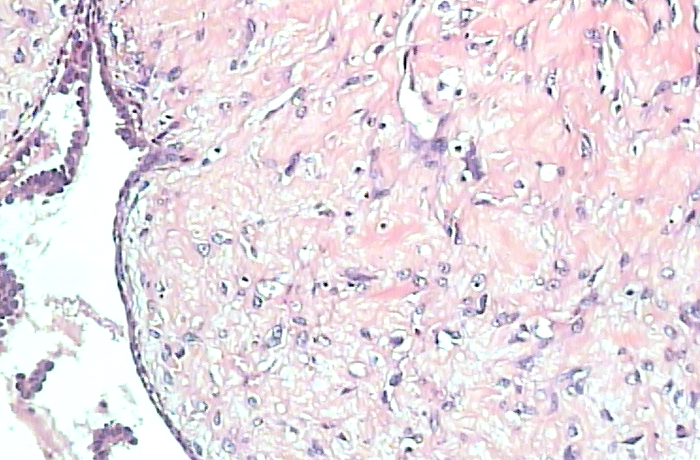}
    \end{minipage}

    \vspace{0.1cm}

    \begin{minipage}{0.12\textwidth}
        \centering
        \textbf{Synthetic data}
    \end{minipage}
    \begin{minipage}{0.2\textwidth}
        \centering
        \includegraphics[width=\linewidth]{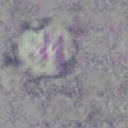}
    \end{minipage}
    \begin{minipage}{0.2\textwidth}
        \centering
        \includegraphics[width=\linewidth]{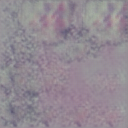}
    \end{minipage}
    \begin{minipage}{0.2\textwidth}
        \centering
        \includegraphics[width=\linewidth]{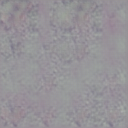}
    \end{minipage}
    \begin{minipage}{0.2\textwidth}
        \centering
        \includegraphics[width=\linewidth]{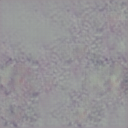}
    \end{minipage}

    \caption{Random samples of real images and synthetic images generated by DCGAN}
    \label{fig:real_synthetic_samples}
\end{figure}

\medskip

\section*{3. Methods}
\subsection*{3.1. Magnification-Invariant Learning via Domain Generalization}
A central challenge in histopathology classification is that image appearance varies substantially across magnification levels, even when the underlying pathology is unchanged [15,16]. To improve cross-magnification generalization, we seek a representation that is highly discriminative for pathology while being minimally informative about magnification. We achieve this using a domain-adversarial learning framework, in which pathology prediction is treated as the primary task and magnification prediction is treated as an adversarial domain task. \\

\noindent Let $x$ denote an input image patch, let $y\in\{0,1\}$ denote the pathology label ($y=0$ benign, $y=1$ malignant), and let $d\in\mathcal{M}$ denote the magnification domain, where
\[
\mathcal{M}=\{40\mathrm{X},\,100\mathrm{X},\,200\mathrm{X},\,400\mathrm{X}\}.
\]
We decompose the model into a feature extractor $f_\theta(\cdot)$, a pathology classifier $g_\phi(\cdot)$, and a magnification-domain classifier $h_\psi(\cdot)$. For each input $x$, we compute the embedding
\begin{equation}
z = f_{\theta}(x),
\label{eq:feature_map}
\end{equation}
the pathology logit $s_y=g_\phi(z)$ and predicted probability $\hat{y}=\sigma(s_y)$,
\begin{equation}
\hat{y} = \sigma\!\left(g_{\phi}(z)\right),
\label{eq:label_head}
\end{equation}
and the magnification logits $s_d=h_\psi(z)$ with $\hat{d}=\mathrm{softmax}(s_d)$.
\begin{equation}
\hat{d} = \mathrm{softmax}\!\left(h_{\psi}(z)\right)
\label{eq:domain_head}
\end{equation}

\noindent The pathology loss is the binary cross-entropy
\begin{equation}
\mathcal{L}_{\mathrm{path}}(\theta,\phi)
=
-\mathbb{E}_{(x,y)}
\left[
y \log \hat{y} + (1-y)\log(1-\hat{y})
\right],
\label{eq:path_loss}
\end{equation}
and the magnification loss is the multiclass cross-entropy
\begin{equation}
\mathcal{L}_{\mathrm{mag}}(\theta,\psi)
=
-\mathbb{E}_{(x,d)}
\left[
\log \hat{d}_{\,\mathrm{enc}(d)}
\right],
\label{eq:domain_loss}
\end{equation}
where $\mathrm{enc}(d)\in\{1,\dots,K\}$ maps magnification to an index and $K=|\mathcal{M}_{\mathrm{train}}|$. Under the LOMO protocol with held-out magnification $m$, we set $\mathcal{M}_{\mathrm{train}}=\mathcal{M}\setminus\{m\}$, so $K=3$.

\noindent We learn magnification-invariant features by solving the domain-adversarial game
\begin{equation}
\min_{\theta,\phi}\;\max_{\psi}\;
\mathcal{L}_{\mathrm{path}}(\theta,\phi)
-\lambda\,\mathcal{L}_{\mathrm{mag}}(\theta,\psi),
\label{eq:minmax}
\end{equation}
where $\lambda\ge 0$ controls the strength of the invariance constraint. Equivalently, optimization can be viewed as alternating updates:
\begin{align}
(\theta,\phi) &\leftarrow \arg\min_{\theta,\phi}\;
\mathcal{L}_{\mathrm{path}}(\theta,\phi)-\lambda\,\mathcal{L}_{\mathrm{mag}}(\theta,\psi),
\label{eq:feature_objective}\\
\psi &\leftarrow \arg\min_{\psi}\;
\mathcal{L}_{\mathrm{mag}}(\theta,\psi).
\label{eq:domain_objective}
\end{align}

\noindent The adversarial interaction is implemented with a gradient reversal layer (GRL). During the forward pass, GRL is the identity; during backpropagation it multiplies the gradient from the domain loss by $-1$ before it reaches the feature extractor. Thus, the domain classifier is trained to minimize $\mathcal{L}_{\mathrm{mag}}$, while the feature extractor is driven to increase $\mathcal{L}_{\mathrm{mag}}$ subject to maintaining low pathology loss, yielding embeddings that are predictive of pathology yet insensitive to magnification.

\subsection*{3.2. Stable Sparse Embedding Signatures}

Beyond predictive accuracy, we seek a compact and transferable representation that summarizes pathology-discriminative information while remaining robust under magnification shift. To this end, we derive a \emph{stable sparse embedding signature} from the learned embedding vectors produced by the feature extractor. Intuitively, the signature identifies a small subset of embedding dimensions that retains most of the discriminative power, enabling interpretability.\\

\noindent Given a trained encoder, we compute an embedding vector for each image patch:
\begin{equation}
d_i = f_\theta(x_i)\in\mathbb{R}^p,\qquad i=1,\dots,n,
\label{eq:embed_extract}
\end{equation}
where $p$ is the embedding dimension. To place all embedding coordinates on a comparable scale, we standardize each dimension using the mean and standard deviation estimated from the training split within each LOMO fold, and apply this same transformation unchanged to the validation and test splits.\\

\noindent Let $D\in\mathbb{R}^{n\times p}$ denote the matrix of standardized embeddings for the training split, and let $y_i\in\{0,1\}$ be the corresponding labels. We fit a sparse logistic model to predict pathology from embeddings:
\begin{equation}
\hat{p}_i = \sigma(\beta_0 + d_i^\top \beta),
\label{eq:sparse_logit}
\end{equation}
where $\beta\in\mathbb{R}^p$ is constrained to be sparse via an $\ell_1$ penalty. Specifically, we estimate $(\beta_0,\beta)$ by minimizing
\begin{equation}
(\hat{\beta}_0,\hat{\beta})
=
\arg\min_{\beta_0,\beta}
\left\{
\frac{1}{n}\sum_{i=1}^{n}
\ell\!\left(y_i,\sigma(\beta_0+d_i^\top \beta)\right)
+\alpha \|\beta\|_1
+\frac{\gamma}{2}\|\beta\|_2^2
\right\},
\label{eq:elastic_net_obj}
\end{equation}
where $\ell(\cdot)$ denotes the Bernoulli negative log-likelihood, $\alpha\ge 0$ controls sparsity, and $\gamma\ge 0$ provides $\ell_2$ stabilization. The resulting \emph{signature} is defined as the support set
\begin{equation}
S = \{j\in\{1,\dots,p\}:\hat{\beta}_j\neq 0\},
\label{eq:signature_support}
\end{equation}
with signature size $|S| \ll p$. \\

\noindent To avoid test-set leakage, the regularization parameters $(\alpha,\gamma)$ are selected using only the validation split within each LOMO fold, optimizing a validation criterion. The final sparse model is then refit on the training split using the selected regularization and evaluated on the held-out magnification test split. \\

\noindent We quantify signature stability across magnification shift by comparing the selected supports $\{S^{(m)}\}$ obtained under each held-out magnification $m\in\mathcal{M}$. In particular, we report (i) per-dimension selection frequencies
\begin{equation}
\pi_j = \frac{1}{|\mathcal{M}|}\sum_{m\in\mathcal{M}} \mathbf{1}\!\left(j\in S^{(m)}\right),
\label{eq:selection_freq}
\end{equation}
and (ii) pairwise support similarity using the Jaccard index
\begin{equation}
J\!\left(S^{(m)},S^{(m')}\right)
=
\frac{|S^{(m)}\cap S^{(m')}|}{|S^{(m)}\cup S^{(m')}|},
\qquad m\neq m'.
\label{eq:jaccard}
\end{equation}
High overlap indicates that the sparse signature captures stable discriminative structure that persists across magnifications, complementing LOMO accuracy metrics with an explicit robustness criterion at the feature-selection level.

\medskip

\section*{4. Results}
We report results under the LOMO protocol to assess robustness to magnification shift and to isolate the contributions of each component of the framework. Comparisons among the baseline, baseline+GAN augmentation, and the proposed domain-general model distinguish gains due to increased training diversity from those due to explicit magnification-invariant learning. 

\subsection*{4.1. Classification Performance under the LOMO protocol}
Table 2 and Figure 4 report holdout-wise classification performance for all three methods under the LOMO protocol. Each magnification is treated as a fully unseen test domain, providing a direct measure of out-of-distribution generalization under magnification shift. The comparison isolates two distinct mechanisms -- distributional augmentation via GAN and representation-level invariance via GRL -- against a supervised baseline, enabling their contributions to cross-magnification robustness to be disentangled. \\

\begin{figure}[H]
    \centering
    \setlength{\fboxsep}{4pt}
    \setlength{\fboxrule}{0.8pt}
    \fbox{%
        \includegraphics[width=0.76\textwidth]{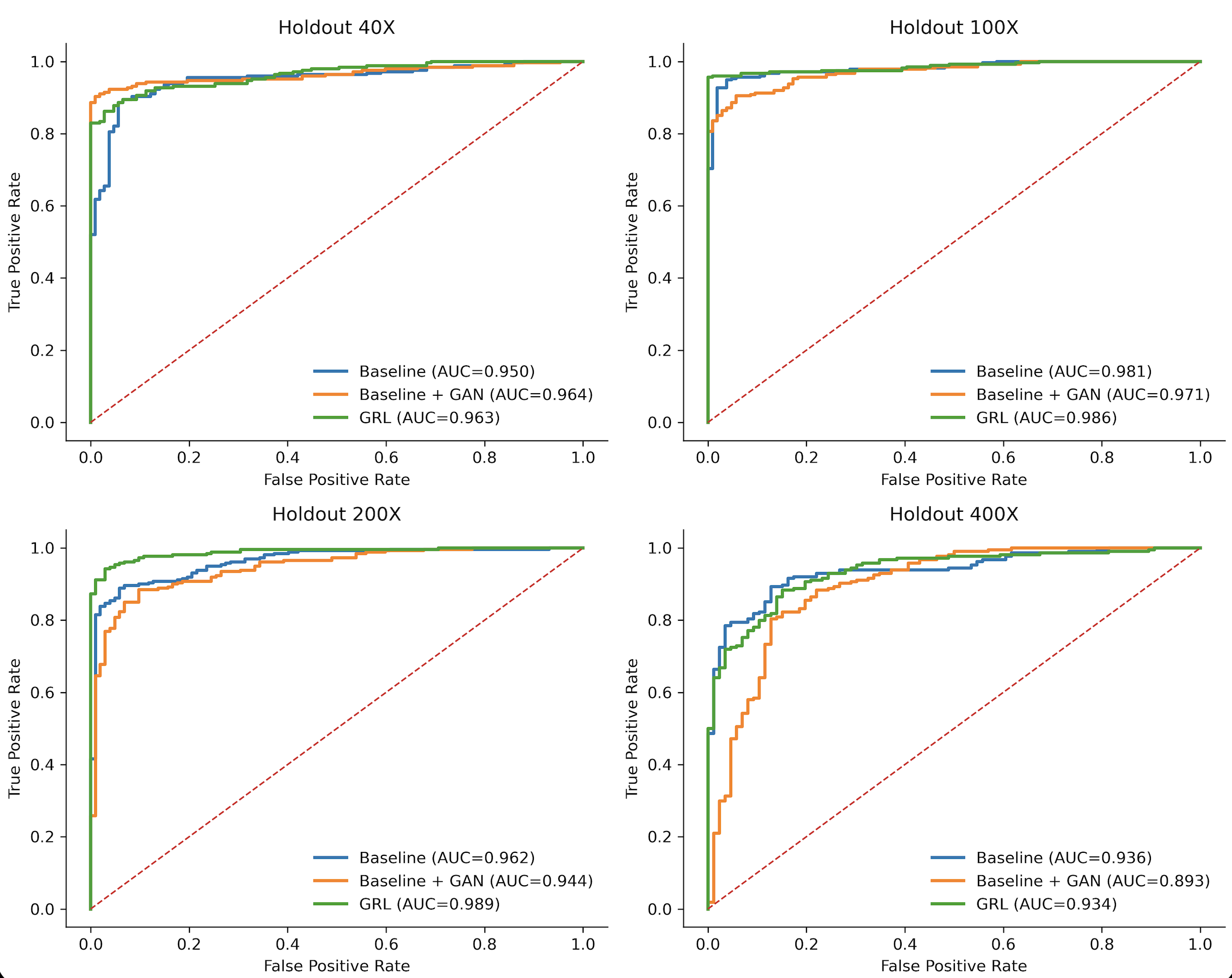}
    }
    \caption{ROC curves across LOMO holdouts for the baseline, baseline+GAN, and GRL models.}
    \label{fig:roc_lomo}
\end{figure}

\noindent No single method dominated across all holdouts, but clear trends emerged. GRL achieved the highest AUC at both 100× (0.986) and 200× (0.989), with its strongest advantage at 200×, where it exceeded the baseline across every reported metric. GAN augmentation performed best when 40× was held out, improving AUC from 0.950 to 0.964 and specificity from 0.654 to 0.701. However, this advantage did not generalize: at 400×, GAN degraded performance substantially relative to both the baseline and GRL, indicating that distributional augmentation without representational constraints introduces instability under harder transfer conditions.

\begin{table}[htbp]
\centering
\begin{threeparttable}
\caption{Summary of all three methods' performance across LOMO holdouts.}
\label{tab:lomo_main_results}
\setlength{\tabcolsep}{5pt}
\begin{tabular}{llccccc}
\toprule
\textbf{Holdout} & \textbf{Method} & \textbf{F1} & \textbf{AUC} & \textbf{Accuracy} & \textbf{Sensitivity} & \textbf{Specificity} \\
\midrule
\multirow{3}{*}{40X}
& Baseline         & 0.909 & 0.95 & 0.867 & 0.959 & 0.654 \\
& Baseline + GAN   & \textbf{0.912} & \textbf{0.964} & \textbf{0.873} & 0.947 & \textbf{0.701}\\
& GRL              & 0.904 & 0.963 & 0.856 & \textbf{0.972} & 0.589 \\
\midrule
\multirow{3}{*}{100X}
& Baseline         & \textbf{0.965} & 0.981 & \textbf{0.95} & 0.945 & \textbf{0.962}\\
& Baseline + GAN   & 0.932 & 0.971 & 0.897 & \textbf{0.971} & 0.702\\
& GRL              & 0.960 & \textbf{0.986} & 0.942 & \textbf{0.971} & 0.865\\
\midrule
\multirow{3}{*}{200X}
& Baseline         & 0.923 & 0.962 & 0.884 & 0.969 & 0.667 \\
& Baseline + GAN   & 0.915 & 0.944 & 0.876 & 0.931 &  0.735\\
& GRL              & \textbf{0.952} & \textbf{0.989} & \textbf{0.928} & \textbf{0.981} & \textbf{0.794}\\
\midrule
\multirow{3}{*}{400X}
& Baseline         & \textbf{0.923} & \textbf{0.936} & \textbf{0.89} & 0.921 &  \textbf{0.814}\\
& Baseline + GAN   & 0.901 & 0.893 & 0.85 & \textbf{0.958} &  0.581\\
& GRL              & 0.919 & 0.934 & 0.88 & \textbf{0.958} & 0.686 \\
\bottomrule
\end{tabular}
\end{threeparttable}
\end{table}

\medskip
\noindent Sensitivity remained uniformly high across all methods and folds, whereas specificity showed the greatest variability. The principal challenge under magnification shift was therefore not the detection of malignant tissue, but the reliable rejection of benign cases. The baseline retained the highest specificity at 100X (0.962) and 400× (0.814), while GRL recovered the largest specificity gains at 200X (0.794 vs 0.667 for the baseline).

\begin{figure}[h]
    \centering
    \includegraphics[width=1\textwidth]{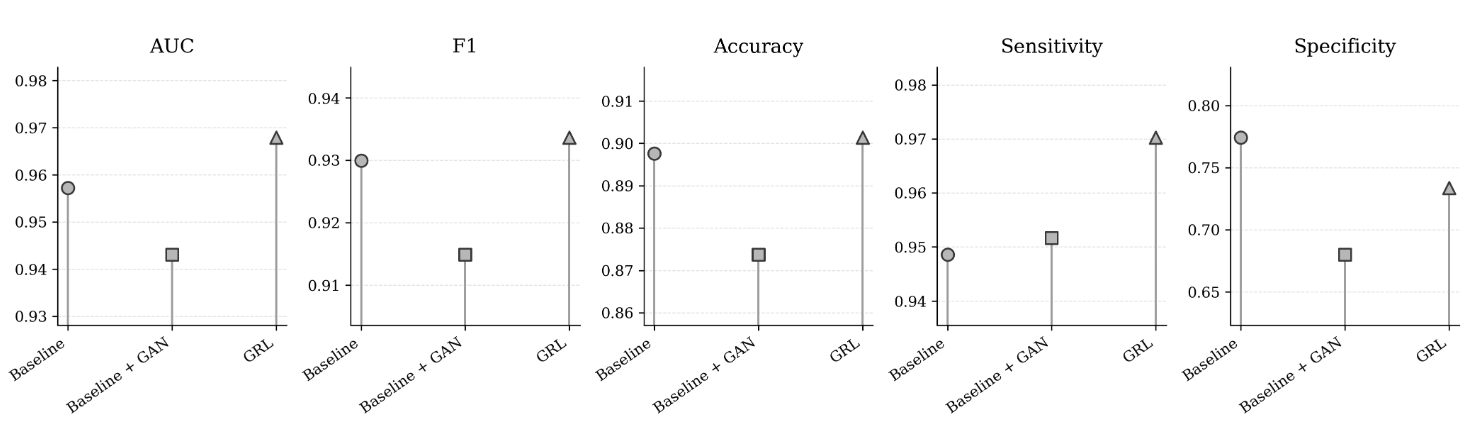} 
    \caption{Performance across LOMO holdouts for all three methods.}
    \label{fig:model}
\label{tab:lomo_main_results}
\end{figure}

\begin{figure}[h]
    \centering
    \includegraphics[width=0.7\textwidth]{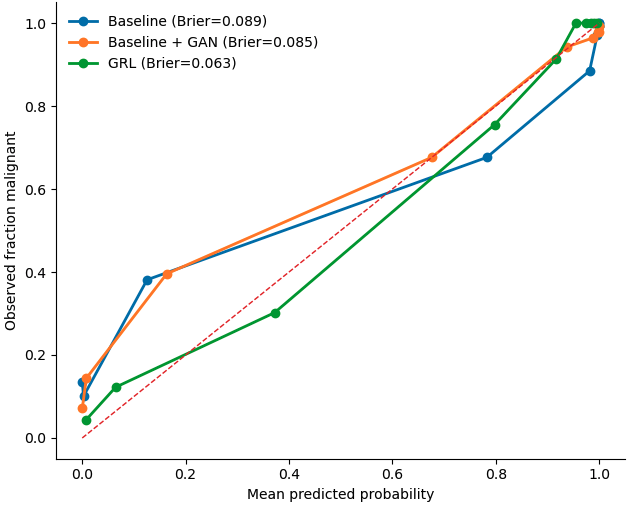} 
    \caption{Calibration on pooled LOMO test predictions for the baseline, baseline+GAN, and GRL models.}
    \label{fig:model}
\end{figure}

\noindent Averaged across holdouts, GRL achieved the strongest performance in AUC, F1-score, accuracy, and sensitivity, while the baseline retained the highest specificity. GAN augmentation did not improve the aggregate performance profile, remaining below both the baseline and GRL in most averaged metrics. Discrimination alone, however, does not establish whether predicted probabilities are reliable. Calibration analysis on pooled LOMO test predictions as seen in Figure 6 shows that the GRL model achieved the lowest Brier score (0.063) compared to the baseline (0.089) and GAN (0.085), indicating that its advantage extended beyond rank-based discrimination to the quality of confidence assignment, a property of direct relevance to clinical decision-making.

\subsection*{4.2. Performance and stability of the sparse embedding signature}
We further examined the learned representations admit a compact sparse signature that remains informative and stable across held-out magnifications. We evaluate whether a small subset of latent dimensions can preserve predictive performance while exhibiting reproducible selection patterns across LOMO folds to ascertain stability, compactness and predictive efficiency.\\

\noindent The Jaccard cross-fold in Figure 7 shows that the GRL improves the reproducibility of the sparse embedding signature under magnification shift. Baseline-derived signatures showed almost no overlap across folds, with pairwise Jaccard indices near zero in all but one isolated comparison. By contrast, GRL produced a more coherent stability pattern, characterized by near-complete overlap between the 100X and 200X folds and moderate overlap with 400X, while 40X remained comparatively distinct. \\

\begin{figure}[H]
    \centering
    \includegraphics[width=0.85\textwidth]{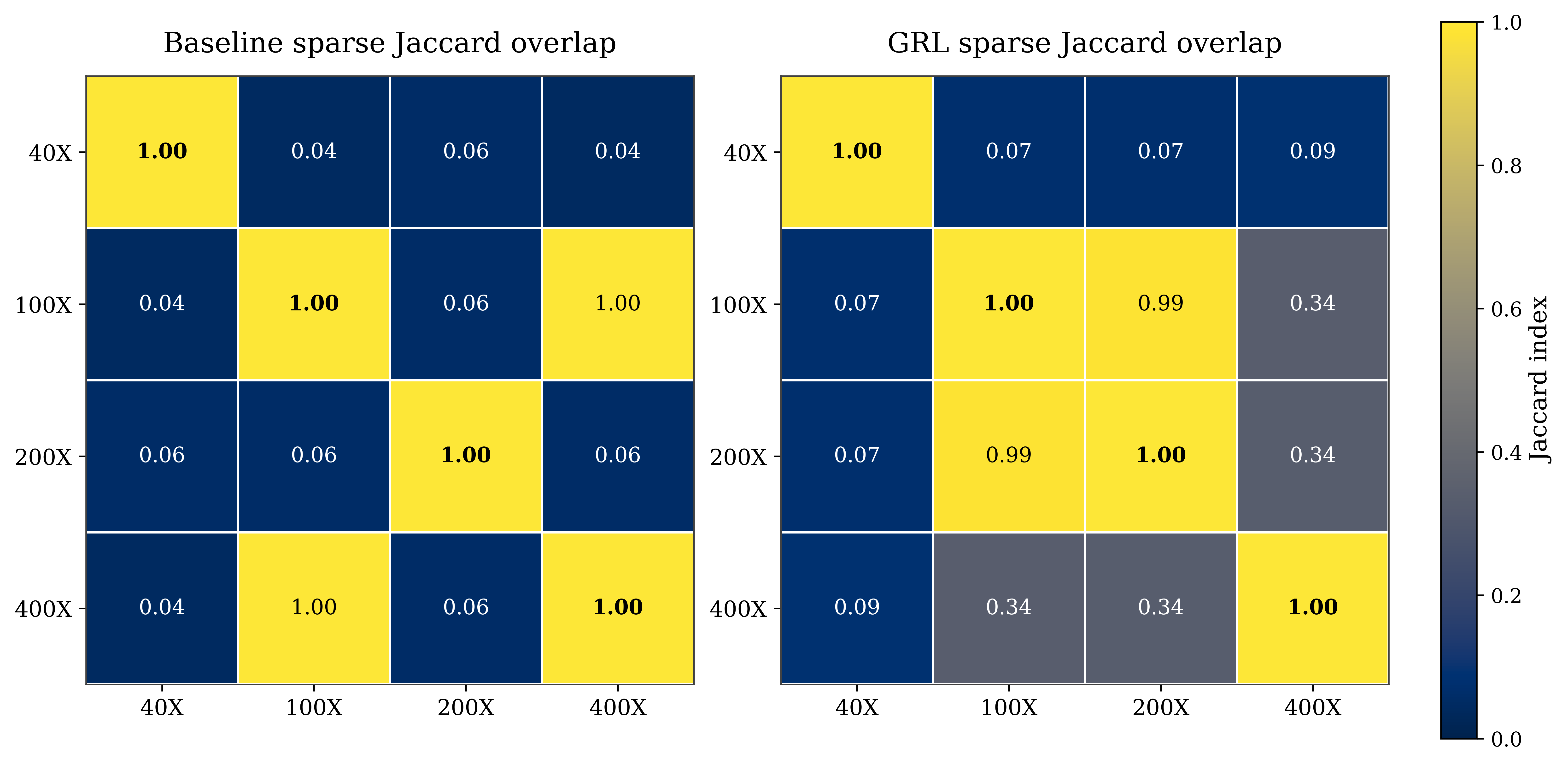} 
    \caption{Pairwise Jaccard overlap of selected sparse dimensions across LOMO holdouts for the baseline and GRL models.}
    \label{fig:model}
\end{figure}

\begin{figure}[h]
    \centering
    \includegraphics[width=0.8\textwidth]{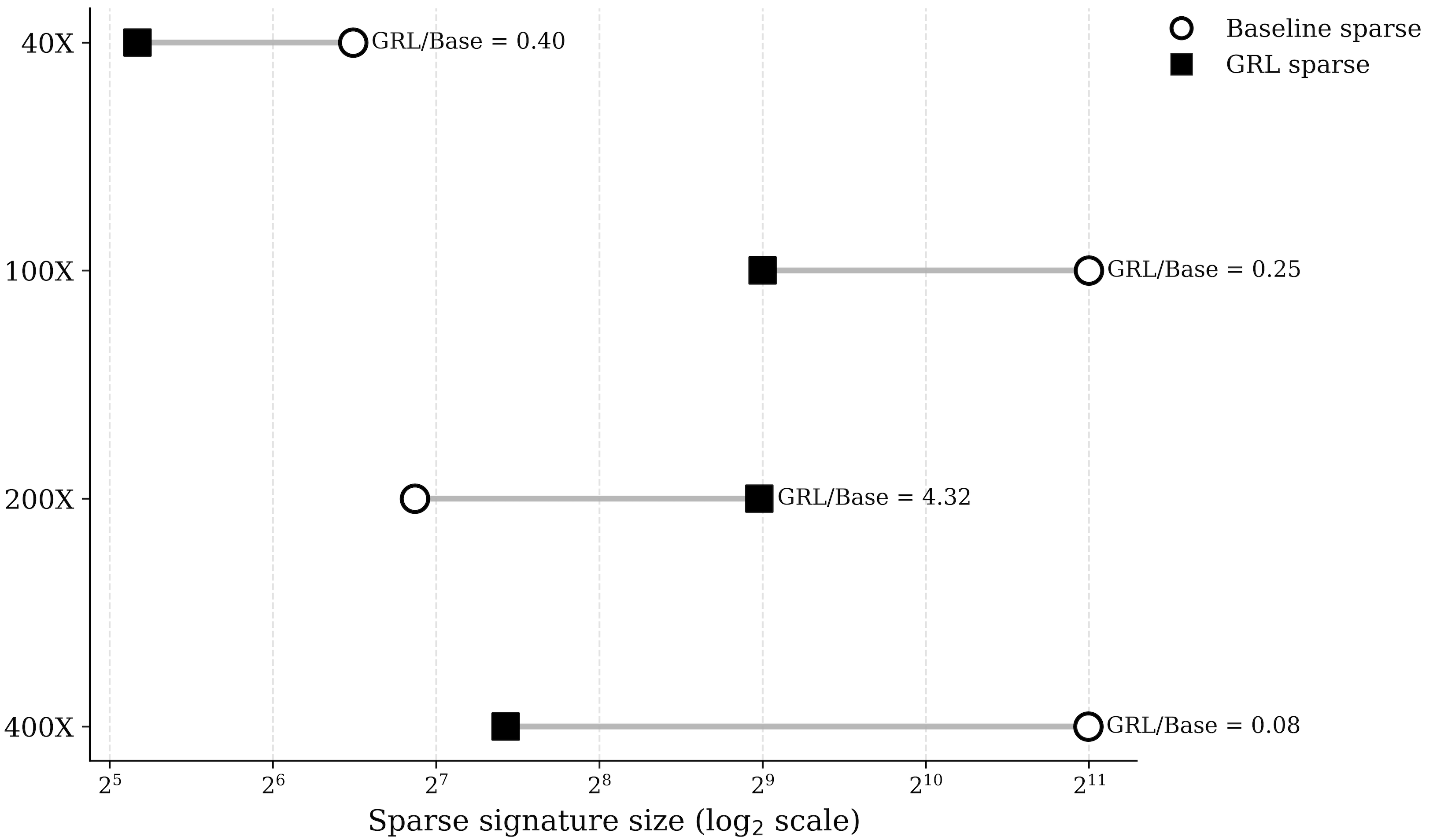} 
    \caption{Sparse signature size across LOMO holdouts for the baseline and GRL models.}
    \label{fig:model}
\end{figure}

\noindent GRL model also produced substantially more compact sparse signatures than the baseline across the majority of holdout magnifications as shown in Figure 8. GRL-to-baseline size ratios were 0.40, 0.25, and 0.08 at 40×, 100×, and 400×, respectively. The 400× ratio represents an order-of-magnitude reduction in signature size. This compressive trend was not universal, however. At 200×, the GRL signature exceeded the baseline by a factor of 4.32, the only condition in which GRL yielded a larger sparse representation. The expanded signature at 200× coincides with the strongest generalization performance across all folds, implying that predictive robustness at this magnification was supported by a broader, more expressive feature set. Signature complexity under GRL thus appears dynamically scaled to the discriminative demands of each holdout condition. \\

\noindent Table 3 indicates that the GRL sparse model was more prediction-efficient than the baseline sparse model. Despite using a markedly smaller signature on average (306.25 versus 1,074.25 dimensions), GRL preserved essentially the same overall AUC (0.967 vs 0.965) and F1-score (0.930 vs 0.931). This pattern suggests that the GRL embedding concentrated discriminative information into a substantially more compact subset of latent features. The corresponding tradeoff was modest: GRL showed slightly lower average accuracy and specificity, but higher sensitivity, consistent with the broader LOMO results in which GRL favored malignant-case detection under magnification shift. \\

\begin{table}[htbp]
\centering
\caption{Comparison of signature size and predictive performance across all LOMO holdouts for Baseline and GRL sparse models}
\label{tab:sparse_holdout_metrics}
\resizebox{\textwidth}{!}{%
\begin{tabular}{llcccccc}
\toprule
\textbf{Method} & \textbf{Holdout} & \textbf{Size} & \textbf{AUC} & \textbf{F1} & \textbf{Accuracy} & \textbf{Sensitivity} & \textbf{Specificity} \\
\midrule
\multirow{5}{*}{Baseline sparse}
& 40X  & 90    & 0.970 & 0.905 & 0.861 & 0.951 & 0.654 \\
& 100X & 2,048 & 0.985 & 0.972 & 0.960 & 0.971 & 0.933 \\
& 200X & 117   & 0.966 & 0.921 & 0.881 & 0.962 & 0.676 \\
& 400X & 2,042 & 0.939 & 0.927 & 0.897 & 0.925 & 0.826 \\
& \textbf{Average} & \textbf{1,074.25} & \textbf{0.965} & \textbf{0.931} & \textbf{0.900} & \textbf{0.952} & \textbf{0.772} \\
\midrule
\multirow{5}{*}{GRL sparse}
& 40X  & 36   & 0.964 & 0.906 & 0.861 & 0.959 & 0.636 \\
& 100X & 512  & 0.985 & 0.953 & 0.931 & 0.971 & 0.827 \\
& 200X & 505  & 0.988 & 0.948 & 0.923 & 0.985 & 0.765 \\
& 400X & 172  & 0.932 & 0.914 & 0.870 & 0.963 & 0.640 \\
& \textbf{Average} & \textbf{306.25} & \textbf{0.967} & \textbf{0.930} & \textbf{0.896} & \textbf{0.969} & \textbf{0.717} \\
\bottomrule
\end{tabular}%
}
\end{table}

\noindent This efficiency pattern was, however, not uniform across holdouts. For 40X and 400X, GRL achieved broadly comparable performance with far smaller signatures, indicating a clear gain in compactness without substantial loss of predictive utility. For 100X, GRL again yielded substantial compression, although with some reduction in threshold-based performance. The 200X holdout showed the opposite behavior: the GRL sparse model required a larger signature, but this was also the setting in which it achieved the clearest predictive advantage.

\section*{5. Conclusion}
Magnification shift degrades histopathology classifiers not because training data is insufficient, but because the learned representation encodes magnification-specific variation that does not transfer. This study shows that constraining the representation, rather than enriching the training distribution, is the decisive factor for robust cross-magnification generalization. Under a strict leave-one-magnification-out protocol, domain-adversarial training via grl consistently outperformed both the supervised baseline and GAN-augmented counterpart in discrimination and calibration, with the clearest margin at 200X. Synthetic augmentation, despite broadening the training distribution, failed to deliver stable gains and actively degraded performance at 400X, a result that cautions against treating data diversity as a proxy for representational robustness. \\

\noindent Beyond classification performance, the sparse embedding analysis uncovered a structural signature of this robustness. GRL embeddings yielded sparse supports that were substantially more reproducible across held-out magnifications, with near-perfect overlap between 100× and 200× folds, alongside a four-fold reduction in average signature size relative to the baseline at matched predictive performance. This compression was not uniform: at 200×, GRL expanded its signature and simultaneously achieved its strongest generalization, a dissociation that reveals signature complexity as an adaptive property of the representation rather than a fixed cost of invariance learning. \\

\noindent These findings suggest a practical point for computational pathology: compact, calibrated, and transferable representations can be learned without more elaborate architectures or large multi-site datasets when the training objective is designed to encourage domain invariance. As histopathology workflows continue to involve heterogeneous acquisition conditions, such representation-level constraints may provide a useful way to improve model robustness at deployment.

\end{document}